# A new transformation for embedded convolutional neural network approach toward real-time servo motor overload fault-detection

Seyed Mohammad Hossein Abedy Nejad[1], Mohammad Amin Behzadi[2], Abdolrahim Taheri[3]

***Abstract***: Overloading in DC servo motors is a major concern in industries, as many companies face the problem of finding expert operators, and also human monitoring may not be an effective solution. Therefore, this paper proposed an embedded Artificial intelligence (AI) approach using a Convolutional Neural Network (CNN) using a new transformation to extract faults from real-time input signals without human interference. Our main purpose is to extract as many as possible features from the input signal to achieve a relaxed dataset that results in an effective but compact network to provide real-time fault detection even in a low-memory microcontroller. Besides, fault detection method a synchronous dual-motor system is also proposed to take action in faulty events. To fulfill this intention, a one-dimensional input signal from the output current of each DC servo motor is monitored and transformed into a 3d stack of data and then the CNN is implemented into the processor to detect any fault corresponding to overloading, finally experimental setup results in 99.9997% accuracy during testing for a model with nearly 8000 parameters. In addition, the proposed dual-motor system could achieve overload reduction and provide a fault-tolerant system and it is shown that this system also takes advantage of less energy consumption.

## I. Introduction

DC servo motors are commonly use in house applications, industries and military systems. They have been chosen for different tasks such as: elevators, cars, trains and etc [1-4]. These tasks might continue for even a year or few seconds. The important point is DC servo motors had to work properly until the given task is done. Any other scenarios may cause damage to the whole system which contains these actuators or even lead to fatality. A common problem that can stop these motors from operating is overloading, this might happen due to environmental issues or commonly higher torque which the motor can overcome, and eventually results in malfunctioning and bad timing.

✉ Abdolrahim Taheri

rahim.taheri@put.ac.ir

Seyed Mohammad Hossein Abedy Nejad

Mohammad.abedy@ut.ac.ir

Mohammad Amin Behzadi

ma.behzadi@mech.sharif.edu

1  Mechanical Engineering Department, Faculty of Engineering, Shiraz University, Shiraz, Iran
2  Mechanical Engineering Department, Faculty of Engineering, Sharif University of Technology, Tehran, Iran
3  Abadan Faculty of Petroleum, Petroleum University of Technology, Abadan, Iran

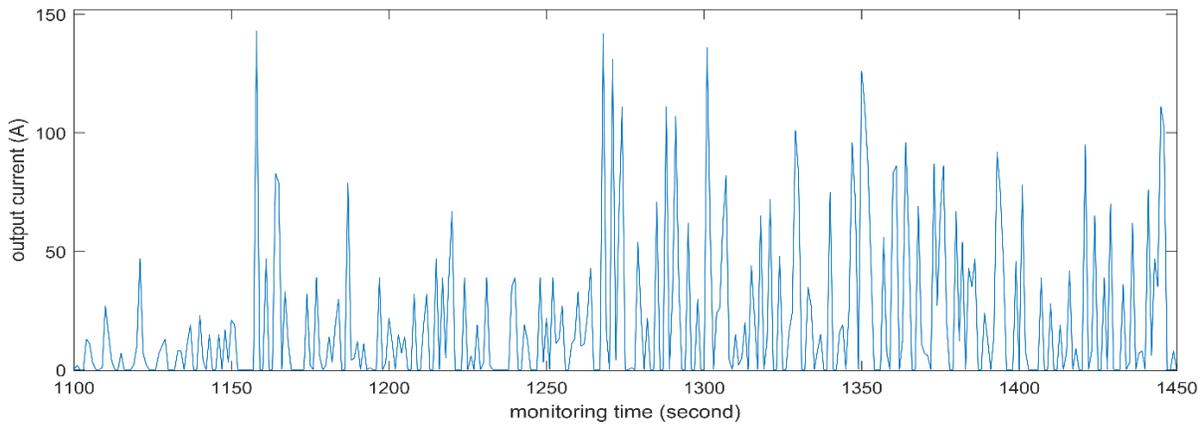

Figure 1. Monitored signal during actuation without fault

Furthermore Available commercial motors in this situation will continue unstoppably to consume as much current as possible to overcome the load, and this will lead to the temperature raise which causes unsafe situations and can lead to damages even to the contiguous equipment [5, 6]. To avoid this, motors operation will be monitored but it's a burden for a human to detect any overload faults distinguish sudden changes, and measure the temperature raise, but this is not the only problem, studies have shown that many countries such as the United States, Hong Kong, and European countries struggle with shortage of skilled operators. Milutin Radonji´c et al at [7], pointed to these problems proposed IOT monitoring, however, this system is cable of enhancement and Therefore a new approach toward automation is required to solve mentioned problems. Thanks to many researchers in the fault-detection, fault -diagnosis, and fault-tolerant, nowadays many tools and solutions are available for the fault-detection. Explicit detection using classic control theory or implicit detection using available methods such as fuzzy logic and adaptive neural Fuzzy inference system (ANFIS), genetic algorithm, Bayesian classifier, support vector machine (SVM), and generally classical machine learning (ML) methods, artificial neural network (ANN), and convolutional neural network (CNN) or generally deep learning methods (DL) [8-12]. In the vast ocean of AI research, Lei Yaguo et al at [13], review ML applications. and Onur Avci, et al [14], review structural fault-detection using classical ML methodologies, these research can be considered a good introduction. In addition Mohammad Sabih and et al in [15] used image classification for fault-detection is similar to our intention. Zhuyun Chen, et al at [16] paper is about mechanical fault extraction using continuous wavelet transform and CNN network provides good insight into signal preprocessing and performing CNN also Lucas C.Brito, et al [17], chooses an unsupervised AI approach to detect faults in rotary equipment. in addition, they implemented Shapley Additive Explanations (SHAP) to interpret classified results from an unsupervised models which is noticeable. Ronny Francis Ribeiro Junior et al [18], carry out signal processing and fault-extraction using one-dimensional input and ANN model. F. Jia, et al at [19] a recommend supervised deep-learning model for fault-detection purposes in presence of a large dataset. Furthermore, it is tangible that Implicit fault detection using AI has many advantages over other methods, to exemplify some of its advantages it will reduce cost remarkably and with many open-source applications and models available through the internet, implementing a new model will be so fast and straightforward.

For our objective two different approaches are available, the first one is unsupervised learning and the other is supervised learning [20-24]. In the first approach, faults are not needed to be identified, but the AI algorithms such as anomaly detection or k-mean clustering can extract faults automatically as a faulty signal has different characteristics, but it is simple to gather an acceptable amount of identified fault and normal signals will be used supervised learning algorithms.

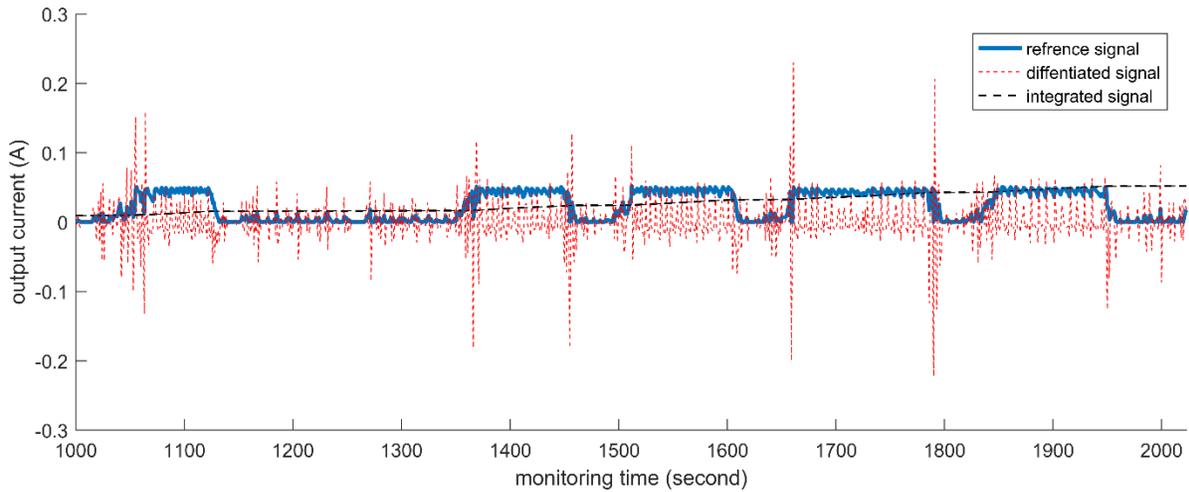

Figure 2. Three channel of data during multi faulty events

also, another reason can be added for our choice although properties like average, standard deviation, and other statistical quantities help to improve our insight over the dataset but the behavior of the fault or technically its average or other properties of even healthy signals are not the same for different task hence it was not possible for us to employed methods such as anomaly detection. A significant subset of supervised learning methods is deep learning, which contains models similar to the neural network, artificial neural network (ANN) is a common model in deep learning, and understanding its functionality and construction is not complex anymore concerning for to available open source tools, however for more complex purposes like computer vision or image processing which input data is a three-dimension or higher order matrix, different neural network architecture is required. Convolutional neural networks (CNN) are state-of-art methods that can detect small to large-scale features from given inputs, for example, they can detect borders, shapes, parts of a human faces, , gender, age, animals, vehicle parts, and so on from given images. It can use to detect faults, but in real-word problems we are mostly expecting signals not images, and signals are one-dimension entities in most cases. So common approach is to build one-dimensional input ANN model to detect faults in monitoring signals, however, ANN models are not the most efficient method at least in our case of overload detection (signal processing) as the nature of signals varies over different speeds and target positions.

Also, sequence models are another available option, for example, Junchuan Shi and et al at [25], perform LSTM sequential networks for planetary gearbox fault-detection, these models receive input through a chain of connected computational units called sequence and data can flow in one or two directions. These models are the best choice for the one-dimensional data processing such as natural language processing and speech recognition, however the complexity and high computational cost which constrain hardware implementation are the drawbacks and prevent us to employed it them for our purpose [26, 27].

Hence in this paper over-loading fault-detection is investigated through by converting the monitored signal to three-dimensional data which whose features can be extracted precisely by implementing a compact CNN model. therefore; an efficient and also mobile model will be obtained with a low computational cost which can be implemented in a low-power microcontroller such as Aurdino. In addition to fault-detection implementation, a synchronous dual-motor system can be used to increase fault-tolerance of the system, it was shown [28] that this system.

takes advantages of reducing energy-consumption this will be discussed in section IV). To present our research rest of the paper is organized as follows, in the next section (0) overloading fault and intuition toward fault detection will be discussed, in the third section (**Error! Reference source not found.**) a compact CNN model is presented. And in the section 0) data acquisition method is described. Successively experimental implementation and results are provided in the section (I). Finally, conclusions are drawn in section VII).

## II. Overloading Faults

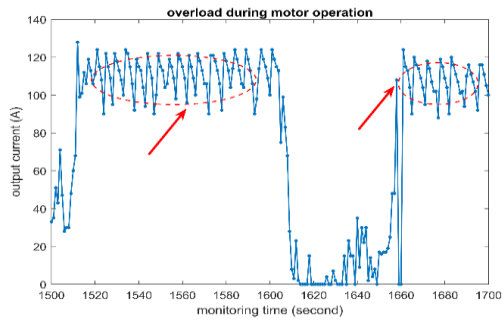

*Figure 3. Faulty signal acquired during monitoring for short range of time*

The first step to establish our fault detection model is to understand overloading faults in monitored signals since labeled data and identified faults is essential for training the model and our supervised approach, hence in this section we inspect collected signals to extract faults. As it is illustrated in Figure 3, on the overloading event output current of the motor will increase and fluctuate over a range until the shutdown command, this pattern will be repeated for the same overloading event, but with different scales over different speeds. The important features to extract are fluctuation range, amount of fluctuation, current speed, target position, the sudden raise slop, the rate of sudden raise in the sampling interval, size of dissipated charges, and the amount of the temperature rise, checking all of these parameters will help an operator to detect faulty signals. The major concern is that many of mentioned parameters must be checked only by monitoring the output signal, this is a burden for the operator besides discerning faulty signals with common fluctuation during actuation which is shown in Figure 1, deteriorating this situation even for a skilled technician. Since it is needed to detect faults for our CNN model, an accurate faulty signal is needed for our dataset, to make this signal interpretable, one idea is to transfer signals to achieve a smooth and interpretable trend.

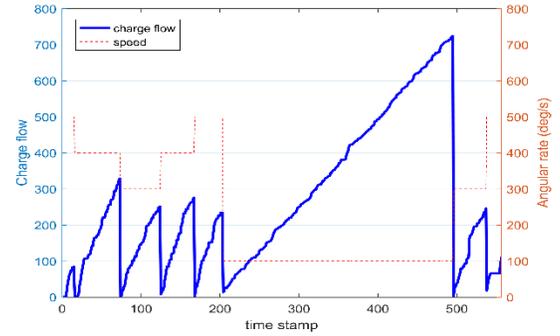

*Figure 4. Faulty signal with integral and differential transform*

Many solutions are available [29, 30], the one which meets our intention is to integrate signals in a specific range of time. As it is shown in Figure 4 the simple integration transformation results in a valuable trend that provide many specifications such as target position or reference speed and also charge flow, but the important point is that the sudden jump in the following trend will be clear and detectable even more than raw signals.

So it could be realized that transformed signals with integration provide even more features to extract faults rather than origin signals and are compatible with our needs, subsequently, these two data can be combined to detect a fault, till here two channels of data for a specific time range are obtained if somehow similar channels of data are obtained, it's possible to construct a three-dimensional array similar to images with RGB channels, therefore, a CNN model can be implemented to extract overloading fault with respect to more details rather than one-dimensional signal. Differentiation is chosen as the third channel to reach the required three-dimensional data. The reason behind this choice is that differentiation would extract changes in the trend remarkably. Now the idea is clarified, the original signal is used in accompaniment of integral and differential transform (Figure 4) to construct our image then trained the CNN model and extract faults from provided image. Finally, it is expected to achieve a dataset that simplify the classification process as we could enhance and augment the data hence compact and low-parameter model can be obtained which helps real-time embedded implementation.

## III. Fault detection CNN-ResNet model

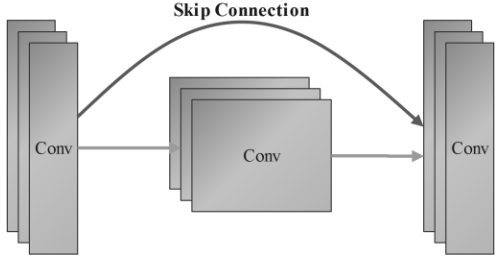

Figure 5. Skip connection example; this connection provide data from previous layers and help to build deeper network

Since a way to obtain three channels from a one-dimensional signal is found , now it's possible to employ CNN [31, 32]. This model can extract and classify data using a pipeline of connected convolutional layers. Each convolutional layer, filters input data with properties like padding[33] and stride [34] and then pass the result through the activation function [35] as well as implements pooling [36]. One of the well-known CNN structures is the ResNet model [37, 38]. This model benefits from skip connections (**Error! Reference source not found.**). This skip connection let produce a deeper but compact network and facilitate information flow through the model. The building block of the ResNet can be expressed as:

$$Y = \mathcal{F}((\omega^{l+2} X^{l+2} + b^{l+2}) + X^l) \quad (1)$$

Where $Y$ is the output of the ResNet Block, respectively $\omega$ and $b$ are the trainable parameter of the network, X is the input given to the block and $l$ represent the block number in the network. The important issue is to determine how to put these convolutional layers together, A compact ResNet model called toy_ResNet is chosen for simplicity and being compact which means fewer parameters but more efficient in comparison to other models. The model summary is given in the Table 1.

Table 1. Model summery of the toy_resnet

| Layer (type) | Param | Connected to |
|---|---|---|
| img (InputLayer) | 0 | - |
| conv2d_7 (Conv2D) | 224 | Img |
| conv2d_8 (Conv2D) | 1168 | conv2d_7 |
| max_pooling2d_1 (MaxPooling2D) | 0 | conv2d_8 |
| conv2d_9 (Conv2D) | 1160 | max_pooling2d_1 |
| conv2d_10 (Conv2D) | 0 | conv2d_9 |
| add_2 (Add) | 1160 | conv2d_10, max_pooling2d_1 |
| conv2d_11 (Conv2D) | 1168 | add_2 |
| conv2d_12 (Conv2D) | 0 | conv2d_11 |
| add_3 (Add) | 0 | conv2d_12, add_2 |
| conv2d_13 (Conv2D) | 1160 | add_3 |
| global_average_pooling2d_1(GlobalAveragePooling2D) | 0 | conv2d_13 |
| dense_2 (Dense) | 576 | global_average_pooling2d_1 |
| dense_3 (Dense) | 130 | dense_2 |
| Total params: 7,914 | | |

## IV. Data acquisition

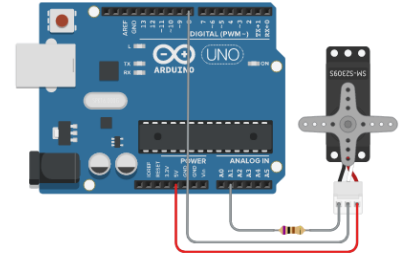

Figure 6. Schematic of monitoring ground current of DC servo motor

For experimental deployment and evaluating the purposed model, first, we need to acquire the CNN model's parameters during training in accompany of validation dataAs Different models may obtain during training and validation, testing data is also considered to choose the best model. To collect data, the output current (ground current) is simply monitored by the analoginput of the microcontroller without the need for any extra sensor. This approach is shown in Figure 6. To understand the procedure the ground voltage of the motor is read by the analog input of the microcontroller in presence of an electrical resistance then using equation (2) output current can be obtained. This data is gathered through discrete time intervals and mapping by default via the microcontroller. This data will be stored in a stack of data with a length of 1024 which reshape and is normalized to a 32*32 array, simultaneously numerical integration and differentiation calculate in a similar 1024-lengthstack of data. For data acquisition purpose, simulated overloading will assert and then each time three stack of data will be stored with a label that express fault, correspondingly three stacks of data for the normal operation of the motor will be stored too. Now the system can train and choose the best model for classification between faulty and normal operation using the CNN model.

$$I = \frac{V}{R} \quad (2)$$

## V. Fault-tolerance and Redundant Dual-motor system

After fault-detection consequent action will fire, for overload purposes, in traditional implementation this action in confined to prompting an alert, however with the dual-motor system is capable of reducing energy consumption intrinsically and consumes even less energy in comparison with a single motor system. this is an advantage besides redundancy. as Redundancy always can help to achieve a fault-tolerant system, a redundant motor can compensate for any fault or failure that occurs during the prime motor. Therefore; in absence of a dimensional restriction in the overall system design and also a limitation of costs, dual-motor is a remarkable implementation for bringing the fault-tolerance ability to the system.

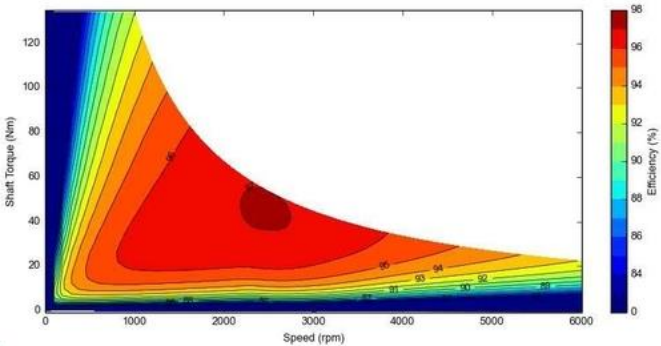

*Figure 7. efficiency map of a servo motor*

In the rest of this section the natural energy-reduction behavior of the dual-motor system will be discussed, it can easily justify the important manner using motor efficiency map, as it is shown in Figure 7. the efficiency of the working motor is dependent on current torque and speed, there is an extreme point that lead to the most efficient operation of the motor, therefore; results in less energy consumption, selecting a motor for any purpose concerning forto this map and this point will lead to optimal working. It is also inferred that working with higher torque will diminish the efficiency of the motor, but if a second motor is added and split the torque this result in better efficiency for the whole system, and its shown in the result section that this can help the system to increase the overall energy-saving rate and achieve energy-consumption less than a single motor system.

a dual-motor system is chosen to accomplish a fault-tolerant system and in addition lower energy consumption.

## VI. Experimental setup and results

As is inferred from Table 1 the CNN model consists of nearly 8000 parameters. Therefore, even typical low-power microcontrollers can afford to store these parameters and then process inputs chunk by chunk. The Chunking data implemented for processing is hired only when the model is implemented into the microcontroller due to hardware limitation, but this also helps to improve calculation faster. As it was discussed our CNN model can be directly implemented into microcontrollers like Arduino, and many free libraries are available such as TensorFlow light or Yolo light. Alternatively, it is possible to use serialized connection to pass and receive data between an external processor which contains the CNN model. Both methods were hired and will be discussed.

In the first step, training and testing data were extracted by the method introduced in the previous section, and eventually, dataset containing 18,000 images, was labeled with 0 as no-fault and 1 as faulty obtained. This dataset was divided to 15,000 images as a training set and respectively 2,000 and 1,000 images as validation and testing.

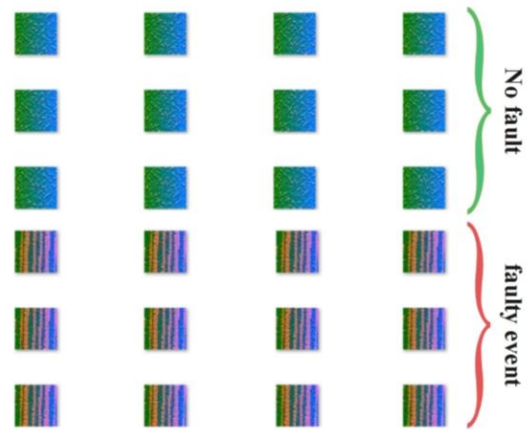

*Figure 8. Extracted images from one-dimensional signals, CNN model will be train by following dataset and will be used to detect faults*

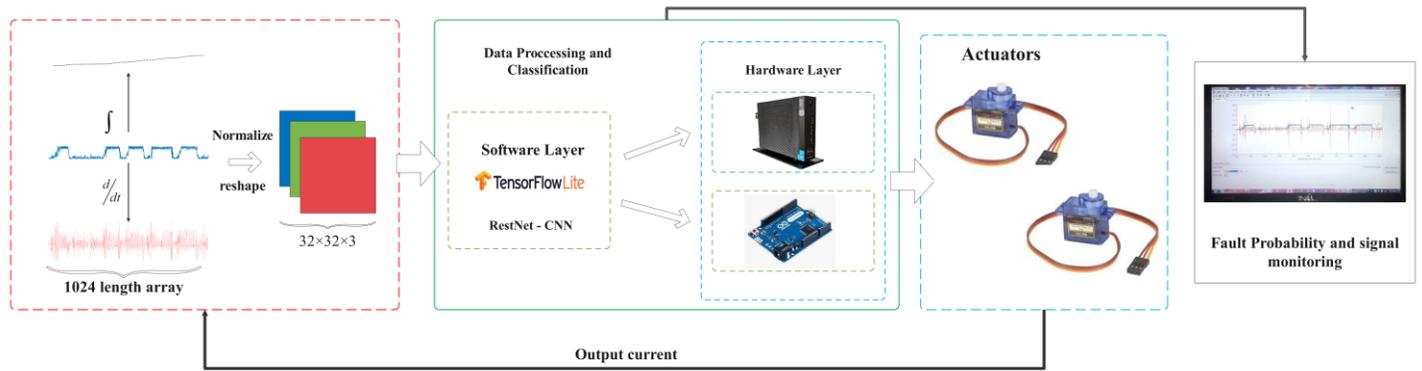

Figure 9. Schematic of proposed procedure aimed at overloading fault-detection

Then the toy-ResNet model is employed to classify our dataset, the model established using Python and Keras framework, the results are provided in Table 2 and Figure 10 and, Figure 11.

Since fault-detection model can be prepared, the next step is to implement the model into hardware layer. Two methods are chosen for implementation:

- Direct implementation (First scenario)

Table 2. Training accuracy and time for ResNet model

| # | ResNet model |
|---|---|
| parameters | 7,914 |
| Time | 37 |
| Accuracy | 0.9995 |

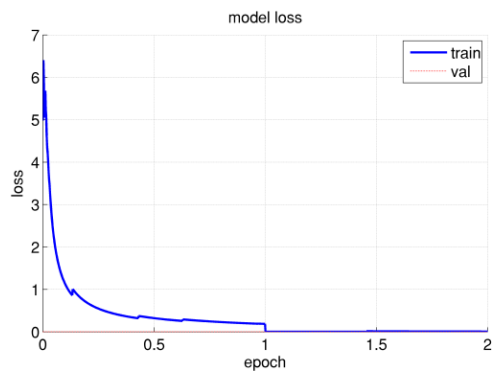

Figure 10. Loss trend of the model training

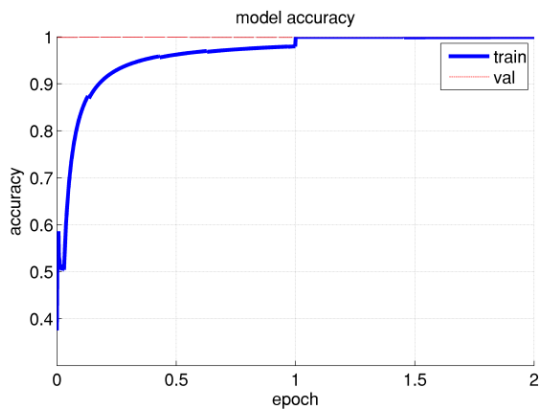

Figure 11. Accuracy trend of the model training

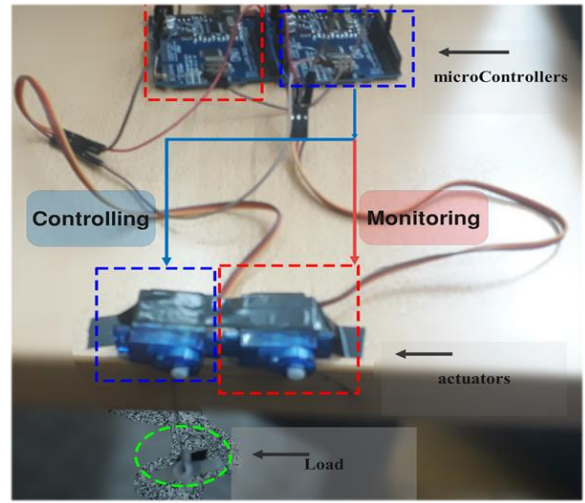

Figure 12. real-time overloading fault-detection via synchronous dual-motor system

In the very first attempt (Figure 12), we implemented a synchronous dual-motor system and inject the CNN model's parameters directly to the Arduino using the tensor flow light library. The proposed system carries two microcontrollers that delegate action to corresponding actuator and in addition monitor the other actuator, this redundancy is helpful in faulty event therefore if one of the motors malfunction the other one can compensate and let the operation continue. As it is discussed that the overall system containing primary and secondary motors will consume less energy in comparison to a single operating motor in the healthy situation this is shown in Figure 14.

- External Processor (Second scenario)

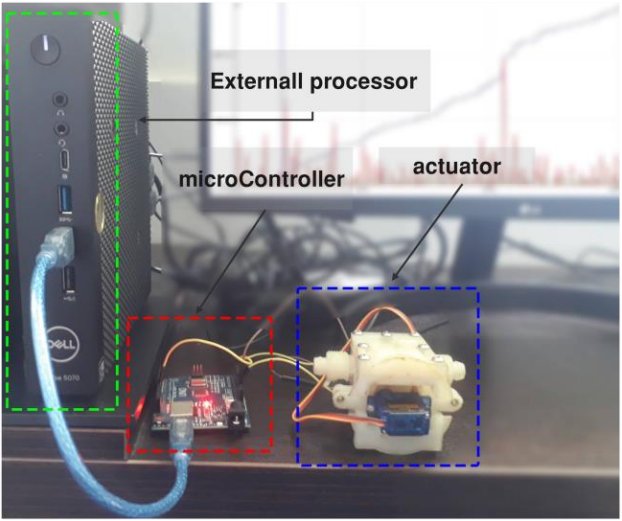

*Figure 13. fault-detection using external processor; in this scenario signals transfer via serialized connection from microcontroller*

The alternative procedure (Figure 13) is to use an external processor which contains our CNN model, this processor receives three channels of data through serialized connection by the microcontroller and determines if overloading is detected.

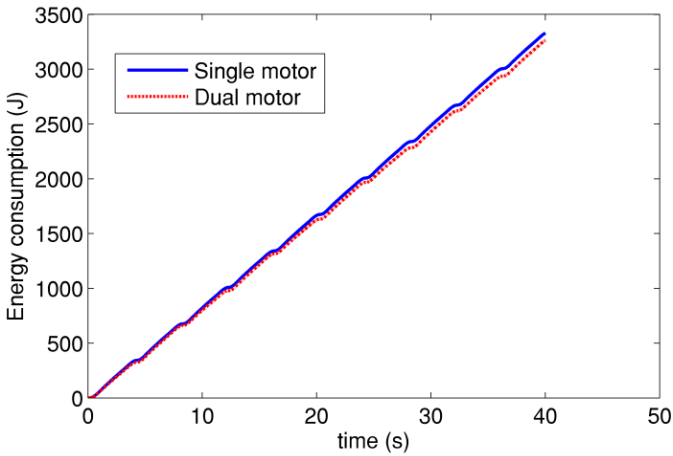

*Figure 14. Energy consumption of singe motor and dual-motor system; about 3% energy is saved.*

## VII. Conclusion

In this research, the feasibility of converting one-dimensional signals to three-dimensional is investigated to access more features therefore simple and shallow models can be employed and implemented at low cost. Overloading fault detection chose as a case study and it was figured out that the output current of DC servo motor can be a good benchmark for overloading fault detection. Then, by integration and differentiation transformation, it was possible to extract three channels similar to the RGB channel of images. In the next step, the Resnet model is hired to classify faulty and normal signals. Finally, the model implemented directly into a low-power microcontroller and into the external processor. Testing was compatible with the expected result.

It was inferred that a much simpler dataset could be built by adding two more dimensions to our data, therefore; a low-parameter model could easily classify the dataset and efficiently detect a fault. To compare the functionality between the direct implementation scenario and the external processor scenario, the first one was a burden to deploy however resulted in faster response since serialized data transportation and response were trimmed. in addition, monitoring could not be established properly. The second scenario was easy to deploy and well monitoring however this implementation caused a delay in the system due to data exchange. Finally, it was concluded that the proposed dual-motor system can reduced energy consumption and help to achieve, a fault-tolerant system. By all observations, using the proposed transformation is recommended which is similar to PID (proportion, integration, differentiation) in control theory to be used in data augmentation and signal transformation. this transformation can be also considered as a special convolutional layer, and in our feature works it was considered to prove that the optimal convolution layer is the same as the proposed PID transformation.